%% file: main.tex
\documentclass{article} 
\pdfoutput=1
\usepackage{ijcai18}
\usepackage{times}
\usepackage[utf8]{inputenc}
\usepackage[small]{caption}
\usepackage{natbib}
\usepackage{graphicx}
\usepackage{color}
\input{preamble}
\title{A Critical Investigation of Deep Reinforcement Learning for Navigation}
\newif\iffinal
\finaltrue

\iffinal
\author{Vikas Dhiman$^1$\footnotemark[1],%
  \, Shurjo Banerjee$^1$\thanks{The first two authors contributed equally.},%
  \, Brent Griffin$^1$,%
  \, Jeffrey M Siskind$^2$,%
  \, Jason J Corso$^1$\\
$^1$EECS, University of Michigan,
Ann Arbor, MI
\and
$^2$ECE, Purdue University,
West Lafayette, IN\\
\texttt{$^1$\{dhiman,shurjo,griffb,jjcorso\}@umich.edu,$^2$qobi@purdue.edu} \\
}
\else
\author{Anonymized}
\fi

\begin{document}
\maketitle
\begin{abstract}
  The navigation problem is classically approached in two steps: an \emph{exploration} step, where map-information about the environment is gathered; and an \emph{exploitation} step, where this information is used to navigate efficiently.
  Deep reinforcement learning (DRL) algorithms, alternatively, approach the problem of navigation in an end-to-end fashion.
  Inspired by the classical approach, we ask whether DRL algorithms are able to inherently explore, gather and exploit map-information over the course of navigation.
  We build upon \citet{MiPaViICLR2017}'s work and 
  introduce a systematic suite of experiments that vary three parameters: the agent's starting location, the agent's target location, and the maze structure.
  We choose evaluation metrics that explicitly measure the algorithm's ability to gather and exploit map-information.
  Our experiments show that when trained and tested on the same maps, the
  algorithm successfully gathers and exploits map-information.
  However, when trained and tested on different sets of maps, the algorithm fails to transfer the ability to gather and exploit map-information to unseen maps.
  Furthermore, we find that when the goal location is randomized and the map is kept static, the algorithm is able to gather and exploit map-information but the exploitation is far from optimal.
  We open-source our experimental suite in the hopes that it serves as a framework for the comparison of future algorithms
  and leads to the discovery of robust alternatives to classical navigation methods.
\end{abstract}

\section{Introduction}
\input{intro-drl-critical-eval}

\section{Related Work}
\input{related-work}

\input{experiments}

\section{Results and Analysis}
\label{sec:analysis}
\input{analysis}

\section{Conclusion and Future Work}
\input{conclusion}

\subsubsection*{Acknowledgments}
%
This work was supported, in part, by the US National Science Foundation under
Grants 1522954-IIS and 1734938-IIS, by the Intelligence Advanced Research
Projects Activity (IARPA) via Department of Interior/Interior Business Center
(DOI/IBC) contract number D17PC00341, and by Siemens Corporation, Corporate
Technology.
Any opinions, findings, views, and conclusions or recommendations expressed in
this material are those of the authors and do not necessarily reflect the
views, official policies, or endorsements, either expressed or implied, of the
sponsors.
The U.S. Government is authorized to reproduce and distribute reprints for
Government purposes, notwithstanding any copyright notation herein.

{\small
\IfFileExists{/z/home/dhiman/wrk/group-bib/shared.bib}{
  \bibliography{/z/home/dhiman/wrk/group-bib/shared,main,main_filtered}
}{
  \bibliography{main,main_filtered}
}
\bibliographystyle{named}
}


\end{document}

%% file: preamble.tex
\usepackage{amsfonts}
\usepackage{amsmath}
\usepackage{amsthm}
\usepackage{booktabs}
\usepackage{comment}
\usepackage{enumerate}
\usepackage{float}
\usepackage{import}
\usepackage[inline]{enumitem}
\usepackage{multirow}
\usepackage{soul}
\usepackage{subfigure}
\usepackage{tikz}
\usepackage{url}
\usepackage{csvsimple}
\usepackage{multicol}
\usepackage{wrapfig}
\usepackage{bbm}

\usetikzlibrary{positioning}

\graphicspath{ {images/} }
\setlist[description]{style=unboxed,leftmargin=2ex}
\tikzset{
every path/.style={line width=0.5pt}
, every node/.style={font=\tiny}
, inputvar/.style={color=red}
, outvar/.style={color=black!50!teal}
, auxvar/.style={color=black!50!olive}
, networknode/.style={inner sep=5.0,rounded corners,fill=black!5}}

\newif\ifblind

\def\act{a}

\def\acttp{\act_{t-1}}

\def\rew{r}

\def\rewtp{\rew_{t-1}}

\def\pos{x}

\newcommand{\LatencyOneGtOne}{Latency-1:\textgreater1}
\newcommand{\NavAiiiCDiDiiL}{NavA3C+D\textsubscript{1}D\textsubscript{2}L}

\newcounter{Benchmark}
\newcounter{BenchmarkB}[Benchmark]
\newcommand{\ditem}[1]{\refstepcounter{Benchmark}\item[\arabic{Benchmark}. {#1}]}

\newcommand{\DistanceInefficiency}{Distance-inefficiency}

%% file: intro-drl-critical-eval.tex
%

Navigation remains a fundamental problem in mobile robotics and
artificial intelligence \citep{SmChIJRR1986,ElCOMPUTER1980}.  The
problem is classically addressed by separating the task of navigation
into two steps: \emph{exploration} and \emph{exploitation}.  In the
exploration stage, an internal \emph{map}-like representation of the
environment is built.  In the exploitation stage, this map is used to \emph{plan a
path} to a given destination based on some optimality criterion (e.g.,
shortest path or minimum energy path).  Despite enjoying
wide success with a variety of environments and sensor types, this
classical approach is heavily dependent on the choice of
feature-representation and map representation given the environment being
navigated. 


\begin{figure}
\includegraphics[width=\columnwidth,trim=0 336pt 2017pt 0,clip]{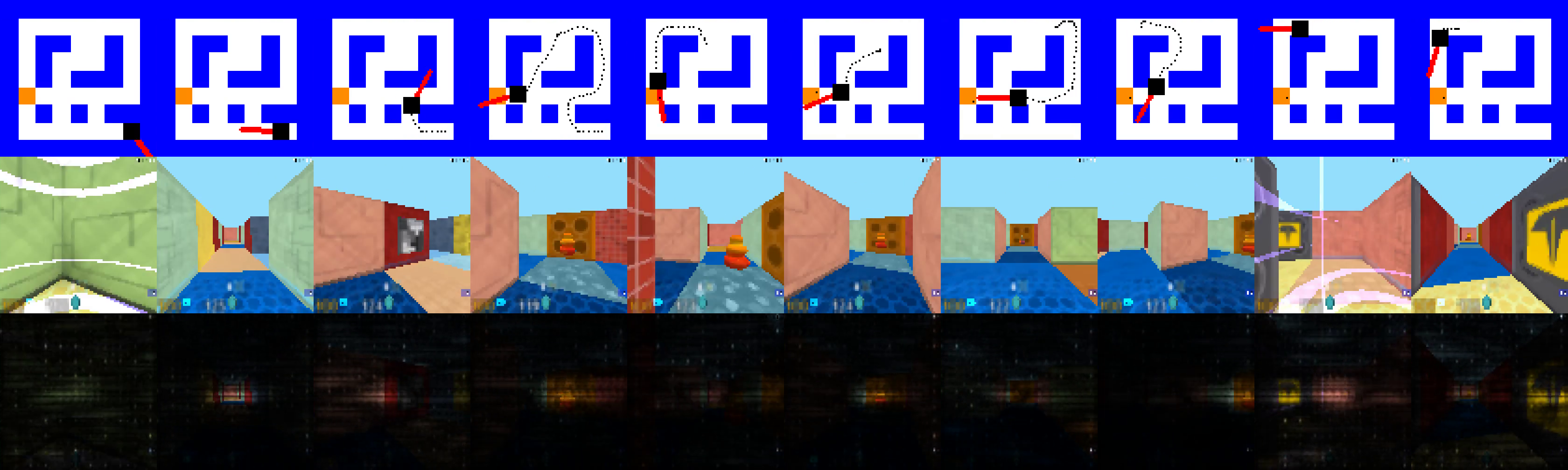}%
\caption{
Snapshots of the path taken by an agent navigating a deepmind lab
    environment.
The top row shows the top view of the robot moving through the maze
    with the goal location marked in orange, the agent marked in black and the
    agent's orientation marked in red. The bottom row shows the first
    person view which  is the
    only input available to the agent besides the previously earned reward.}
\label{fig:training-qualitative}
\end{figure}%
%
%

Recently, end-to-end navigation methods have gained traction, bolstered
by advances in Deep Reinforcement Learning (DRL)
\citep{MnBaMiICML2016,SiHuMaNATURE2016,LePaKrISER2017,MiPaViICLR2017,OhChSiICML2016}.
In these methods, the manual design of the input features,
intermediate map-building and
path-planning stages is replaced with the design of a \emph{reward
function}, a one-dimensional proxy for the navigation objective.  DRL
agents then learn intermediate map representations tailored to the
maximization of this reward function. 

The potential for such simpler formulations of the navigation problem is
rich. For example, when the reward function is customized to the finding
of destinations within a maze, the resulting trained agents may discover
features and map representations that allow for more efficient completion of the navigation
task than human-interpretable maps.

Recent works in the field, however, fail to measure the algorithm's ability to gather and exploit map-information \citep{ZhMoKoICRA2017,KuSaGaAPA2016,jaderberg2016reinforcement,gupta2017cognitive}.
Agents are
usually tasked with finding a destination, sometimes in an
unseen environment and evaluated on some measure of time taken to find the destination.
Though agents may learn to remember map
information to find the destination faster as they navigate, the evaluation
metrics used by these works do not explicitly separate and measure the
agent's facilities for exploration and exploitation.  We believe that
such an explicit separation is a necessary step for the formulation of
agents that perform well on the meta-task of navigation and can someday
provide robust alternatives to classical methods. 

Work by \citet{MiPaViICLR2017} represents a first step in this direction.
Their experimental setup consists of an agent randomly spawned
within an environment whose map-structure remains constant throughout training and
testing.  The agent is tasked with discovering a goal location within
the environment. Upon finding the goal, it is rewarded and randomly
reinitialized. Episodes consist of fixed time-length intervals. To
maximize reward, agents must first find the goal and then repeatedly
revisit it during the course of an episode. The task is thus explicitly
oriented towards the creation and study of agents that must perform both
exploration and exploitation to maximize reward. To measure these
effects, the authors report the \emph{\LatencyOneGtOne{}} metric, the
ratio of time taken to first find the goal to the
average time taken to revisit the goal. 

This work builds upon the experimental setup introduced in their work by
further seeking to understand exactly where DRL-based navigation methods
succeed and fail across a variety of static and random environments.
We train and evaluate the algorithm 
on problems of increasing difficulty. In the easiest
stage, we keep the spawn location, the goal location, and the map constant over
the training and testing. To increase the difficulty, we incrementally
randomize the spawn location, goal location and map structure until
all three variables are random.  In the most difficult stage, agents are trained
on 1000 random maps and tested on 100 previously unseen maps. In
addition to reporting the \emph{\LatencyOneGtOne{}} metric, we introduce
the \emph{\DistanceInefficiency{}} metric, the ratio of distance
covered by the agent as compared to the approximate shortest path length to the
goal.

In the case where environments are kept static throughout training and
testing, we find that agents are able to perform effective exploitation
by consistently finding the goals faster post-goal discovery
(\emph{\LatencyOneGtOne{}} is greater than 1), in line with the
findings of \citet{MiPaViICLR2017}. Furthermore, of these cases
where the goals are static, the \emph{\DistanceInefficiency{}} is
approximately 1, indicating near-optimal goal discovery; by contrast,
cases with randomized goals perform sub-optimally. For the cases where
the agents are trained on 1000 maps and tested on 100 unseen maps, we
find no evidence that map gathering and exploitation is taking place.
To further
highlight this, we qualitatively assess the navigational strategies
utilized by these agents across simplified \emph{planning maps} that
require the agents to make simple binary decisions at junction points.
Finally, we compute attention maps to qualitatively analyze the
environmental cues utilized by these agents to make decisions at
different points in the map. 

To summarize, our contribution is two fold.
First, we propose a systematic suite of experiments along with a set evaluation metrics that explicitly evaluate the ability of algorithms to exploit map information.
Second, we answer whether a representative DRL algorithm is able to exploit map information under this experimental suite.

%% file: related-work.tex
\paragraph{Localization and mapping}
Localization and mapping for navigation is a classic problem in mobile robotics and sensing.
\citet{SmChIJRR1986} introduced the idea of propagating spatial uncertainty for robot localization while mapping, and \citet{ElCOMPUTER1980} popularized Occupancy Grids.
In the three decades since this seminal work, the field has exploded with hundreds of algorithms for many types of sensors (e.g., cameras, laser scanners, sonars and depth sensors).
These algorithms vary in the amount of detail they capture in their respective maps. For example, topological maps, like \citet{KuCOGSCI1978}, aim to capture as little information as possible while occupancy grid maps such as \citet{ElCOMPUTER1980}, aim to capture metrically accurate maps in resolutions dependent upon the navigation task.

All these approaches require significant hand-tuning for the environment, sensor types, and navigation constraints of the hardware.
In contrast, end-to-end navigation algorithms optimize the detail of map storage based on the navigation task at hand. This significant advantage makes end-to-end navigation ripe for exploring.

\paragraph{Deep reinforcement learning}
DRL gained prominence recently when used by \citet{MnKaSiNIPSDLW2013,MnKaSiNATURE2015} to train agents that outperform humans on Atari games; agents that were trained using only the games' visual output.
More recently, DRL has been applied to end-to-end navigation \citep{OhChSiICML2016,MiPaViICLR2017,ChLaSaNIPS2016}.
It is common for agents to be trained and tested on the same maps with the only variation being the agent's initial spawn point and the map's goal location \citep{MiPaViICLR2017,ZhMoKoICRA2017,KuSaGaAPA2016}. 
Even when the agents are tested on unseen environments \citep{MnKaSiNATURE2015,jaderberg2016reinforcement,gupta2017cognitive}, they are evaluated using metrics which only measure the exploration abilities of the agents.
By exploration abilities, we mean that the agent is tasked to find a goal location or object in the unseen environment, and the reported metrics are some variation of the time taken to find the goal.
On the other hand, a metric for exploitation ability will measure the ability of the agent to find a visited location in the unseen map again.

\citet{OhChSiICML2016} test their algorithm on random unseen maps, but their agents are trained to choose between multiple potential goal locations based on past observations.
The episodes end when the agent collects the goal, so there is no requirement for the algorithm to store map information during exploration.
Thus, their agents decide to avoid a goal of a particular color while seeking other colors rather than remembering the path to the goal.
Similarly \citet{parisotto2017neural} extend \citet{OhChSiICML2016} to work for long time ranges by indexing the memory with spatial coordinates.
\citet{ChLaSaNIPS2016} test their method on unseen maps in the VizDoom environment, but only vary the maps with unseen textures. Thus, their agents are texture invariant, but train and test on maps with the same geometric structure.

%


In this work, we propose an extension to the study of these methods in a more comprehensive set of experiments to address the question of whether DRL-based agents remember enough information to obviate mapping algorithms or may in fact need to be augmented with mapping for further progress.

%% file: experiments.tex
\section{Environmental Setup}
We evaluate the algorithms on a simulated environment. 
We use the same game engine as \citet{MiPaViICLR2017},
called Deepmind Lab \citep{BeLeTeARXIV2016}.
The game is setup so that an agent is placed within a randomly
generated maze containing a \emph{goal} at a particular location,
$\pos_g$.  On hitting the goal, the agent \emph{re-spawns} within the
same maze while the goal location remains unchanged.  The maze is
scattered with randomly placed smaller apple rewards (+1) to encourage
initial explorations and the goal is assigned a reward of $\rew_g = +10$.
The agent is tasked with finding the goal as many times as possible
within a fixed amount of time ($T=1200$ steps for our experiments),
re-spawning within the maze, either statically or randomly, each time it
reaches the goal. We include a small wall penalty (-0.2) that pushes the
agent away from the wall to prevent agents from moving along the walls
during initial explorations. At every point the agent must choose
between moving forward or backward or rotating left or right. Following
\citet{ChLaSaNIPS2016}, we use randomly textured walls in our mazes so
that the policies learned are texture-independent.

We generate 1100 random maps using recursive depth-first search-based
maze generation methods \citep{Aycock2016}.  Of the first 1000 maps, 10
are randomly selected for our static-map experiments. For our unseen map
experiments, agents are trained on increasing subsets of the first 1000
maps and tested on the remaining 100.

\section{Network architecture}
\input{fig-nav-a3c-horiz-figure}
Our network architecture is a simplified version of the
\NavAiiiCDiDiiL{} model used by \citet{MiPaViICLR2017}.  A schematic of
the architecture is shown in Fig~\ref{fig:architectures}.  
We chose \NavAiiiCDiDiiL{} as a representative DRL architecture.
This architecture with a CNN as an encoder and RNN on the top is a vanilla DRL architecture that has shown promise in multiple problem domains \citep{hausknecht2015deep}.

At every time-step the architecture is fed three inputs: the current
image $I_t$, the previous action $\acttp$, and the previous reward $\rewtp$ and
is tasked with estimating the value function and policy at every point.
Similar to their setup, we utilize auxiliary outputs of the loop closure
signal $L$ and predicted depth $D_1$ and $D_2$ to quicken training.

We use the Asynchronous advantage actor-critic (A3C) algorithm
\citep{MnBaMiICML2016} to perform reinforcement learning.  A3C
falls in the category of policy gradient algorithms that works by
jointly optimizing the policy function and the value function via the
estimation of their gradients.

\input{fig-latency-goal-reward}
\input{evaluation_metrics}

\input{intro-drl-nav-challenge}

%% file: fig-nav-a3c-horiz-figure.tex
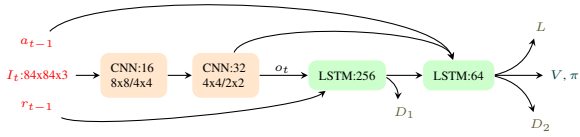
\begin{figure}%
\begin{center}
\scalebox{0.8}{\input{fig-nav-a3c}}%
\end{center}
\caption{
Modified \NavAiiiCDiDiiL{} architecture.
The architecture has three inputs: the current image $I_t$, the previous action $\acttp$, and the previous reward $\rewtp$.
As shown by \protect\citet{MiPaViICLR2017}, the architecture improves upon vanilla A3C architecture by using auxiliary outputs of loop-closure signal $L$ and predicted depth $D_1$ and $D_2$.
}
\label{fig:architectures}
\end{figure}

%% file: fig-nav-a3c.tex
\usetikzlibrary{positioning}
\begin{tikzpicture}
\node[networknode,fill=green!20] at (10.5, 2.5) (lstm2) {LSTM:64};
\node[networknode,fill=green!20,left=0.6 of lstm2] (lstm) {LSTM:256};
\node[networknode,fill=orange!20,left=0.8 of lstm]  (enc2) {\parbox{8ex}{CNN:32\\4x4/2x2}};
\node[networknode,,fill=orange!20,left=0.4 of enc2]  (enc) {\parbox{8ex}{CNN:16\\8x8/4x4}};
\node[left=0.4 of enc, inputvar] (It) {$I_t$:84x84x3};
\node[above=0.1 of It,inputvar] (at) {$\acttp$};
\node[below=0.1 of It,inputvar]  (rt) {$\rewtp$};
\node [right=0.8 of lstm2,outvar] (V)  {$V$, $\pi$};
\node [shift={(-0.4,0)},above=0.4 of V,auxvar] (pi)  {$L$};
\node [shift={(-0.4,0)},below=0.4 of V,auxvar] (D)  {$D_2$};
\draw [-stealth] (It) edge (enc);
\draw [-stealth] (at) edge [in=110,looseness=0.4] (lstm2);
\draw [-stealth] (enc2) edge [bend left=70,looseness=0.4] (lstm2);
\draw [-stealth] (enc) edge (enc2);
\draw [-stealth] (enc2) edge node [shift={(0.225,0.1)},left]{$o_t$} (lstm);
\draw [-stealth] (rt) edge  [bend right,looseness=0.3] (lstm);
\draw [-stealth] (lstm) edge node[below=0.4,auxvar] (D1) {$D_1$} (lstm2);
\draw [-stealth] (lstm) edge [bend left] (D1);
\draw [-stealth] (lstm2) edge [bend right] (pi);
\draw [-stealth] (lstm2) edge  (V);
\draw [-stealth] (lstm2) edge [bend left] (D);
\end{tikzpicture}

%% file: fig-latency-goal-reward.tex
\begin{figure*}[h]%
  \includegraphics[width=\linewidth]{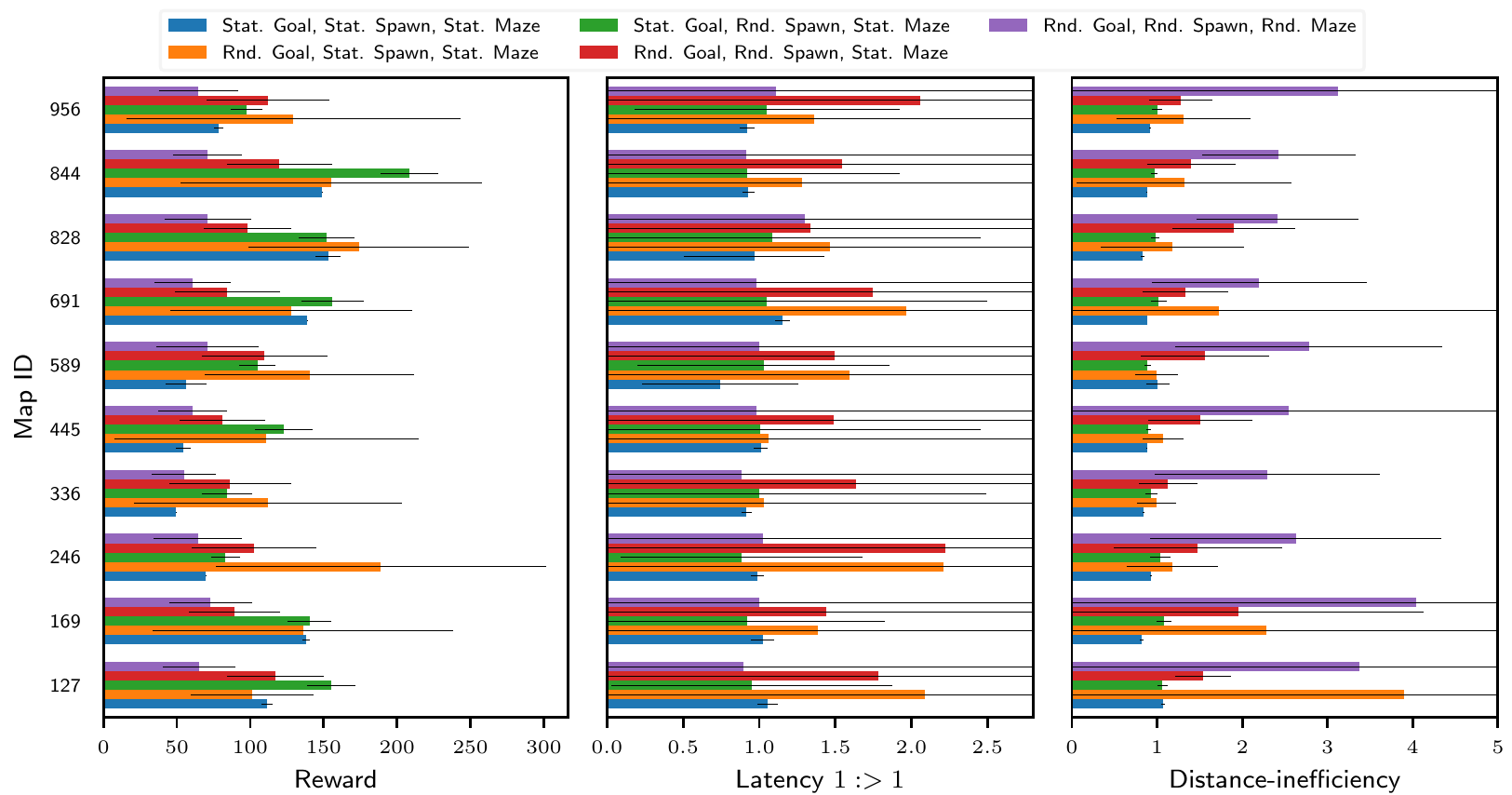}
  \caption{
    We evaluate the algorithm on ten randomly chosen maps with problems of increasing difficulty as described in Sec.~\ref{sec:navtasks}.
    The vertical axis shows IDs of the ten maps from the randomly
    generated 1100 maps on which experimental results are repeated.
    We note that when the goal is static, the rewards are consistently
    higher as compared to random goals.
    With static goals, the metric \DistanceInefficiency{} is close to 1,
    indicating that the algorithm is able to find the shortest path.
    However, with random goals, the agent struggles to find the shortest path.
    From the \LatencyOneGtOne{} results we note that the algorithm does
    well when trained and tested on the same map but fails to generalize
    to new maps when evaluated on the ability to exploit the information
    about the  goal location.
    Note that the \LatencyOneGtOne{} metric for cases of static goals
    is expected to be close to one because the location of goal is
    learned at training time.
    The figure is best viewed in color.
  }%
\label{fig:latency-goal-reward}%
\end{figure*}

%% file: evaluation_metrics.tex
\section{Evaluation Metrics}

In addition to reporting reward, we evaluate the algorithms in 
terms of the \emph{\LatencyOneGtOne{}} and the \emph{\DistanceInefficiency{}} metrics.

Following \citet{MiPaViICLR2017}, we report \emph{\LatencyOneGtOne{}}, a
ratio of the time taken to hit the goal for the first time (exploration
time) versus the average amount of time taken to hit goal subsequently
(exploitation time). To define this metric more concretely, 
let the position of agent at any time $t$ be $\pos_t$.
Let $\tau_{1:N} = \{ t \in [0, T] \mid \|\pos_t - \pos_g\| < \epsilon \}$, be the ordered set of times when the agent hits the goal during a episode, where $N$ is the number of times agent hits the goal and $\epsilon$ is a distance threshold on hitting the goal.
Let $\tau_i$ denote the time step when the agent hits the goal $i$\textsuperscript{th} time.
\begin{align*}
  \text{\LatencyOneGtOne} &= 
  \frac{(N-1) \tau_1}{\tau_{N} - \tau_1} & \text{if }  N >= 2
\end{align*}%

This metric is a measure of how efficiently the agent exploits map information to find a shorter path once the goal location is known. 
If this ratio is greater than 1, the agent is doing better than random exploration and the higher the value, the better its map-exploitation ability.
Note that the metric is meaningful only when the goal location is unknown at evaluation time.

\emph{\DistanceInefficiency{}} is defined to be the ratio of the total
distance traveled by the agent versus the sum of approximate shortest
distances to the goal from each spawn point. The metric also disregards
goals found during exploration times as the agent must first find the
goal before it can traverse the shortest path to it.  Note that the
shortest distance between the spawn and goal locations is computed as a
manhattan-distance in the top-down block world perspective and hence is
an approximation. 
Mathematically, 
\begin{align*}
\text{Dist-ineff.} &=
    \frac{ \sum_{i=1}^{N-1} \sum_{t=\tau_i + 1}^{\tau_{i+1} - 1} \|\pos_{t+1} - \pos_{t}\| }
         { \sum_{i=1}^{N-1} p(\pos_{\tau_i + 1}, \pos_g) } \enspace, & \text{if } N >= 1
\end{align*}%
where $p(\pos_{\tau_i +1}, x_g)$ denotes the approximate shortest path distance between spawn location $\pos_{\tau_i+1}$ and goal location $\pos_g$.

While the \LatencyOneGtOne{} measures the factor by which subsequent
paths to the goal post-goal finding is shorter than the initial
exploration path, the \DistanceInefficiency{} measures the length of
this path with respect to the shortest possible path. 

We report all the metrics averaged over 100 episodes of 10 randomly
chosen maps in Fig~\ref{fig:latency-goal-reward}.

%% file: intro-drl-nav-challenge.tex
\subsection{Experiments}
\label{sec:navtasks}
We evaluate the \NavAiiiCDiDiiL{} algorithm on maps with 5 stages of difficulty, by controlling the randomness of three basic parameters: the spawn location, the goal location, and the map structure.

\begin{description}
  \item[Spawn location] We consider cases where the spawn location is \emph{static} in all the training and testing episodes and when the spawn location is \emph{randomized} for every spawn.
  Note that the agent re-spawns every time it hits the goal.
  \item[Goal location] The goal location is \emph{static} when it is fixed for all the training and testing episodes.
  In the random case it is \emph{randomized} for every new episode.
  \item[Map structure] In the \emph{static map} case, we train and test on the same map.
  We perform static map experiments (training and testing) on ten random maps.
  In the \emph{random map} case, we sample a map from 1000 maps for every episode and test on the ten maps chosen for the static map experiments.
  In Section~\ref{sec:effect-apples-texture}, we evaluate the algorithm on a disjoint set of 100 test maps.
\end{description}

We investigate variations of randomness of the above parameters and propose these experiments as a 5-stage benchmark for all end-to-end navigation algorithms.

\setcounter{Benchmark}{0}
\begin{description}
  \ditem{Static goal, static spawn, and static map}
  \label{prob:sss}
  To perform optimally on this experiment, the agent needs to find and learn the shortest path at training time and repeat it during testing. 

  \ditem{Static goal, random spawn and static map}
  This is a textbook version of the reinforcement learning problem, especially in grid-world \cite{SuBaBOOK1998}, with the only difference being that the environment is partially observable instead of fully observable.
  This problem is more difficult than Problem~\ref{prob:sss} because the agent
  must find an optimal policy to the goal from each possible starting point in the maze.
  \ditem{Random goal, static spawn, and static map}
  The agent can perform well on this experiment by remembering the goal location after it has been discovered and exploiting the information to revisit the goal faster.  
  
  \ditem{Random goal, random spawn, and static map}
   To perform optimally, the agent must localize itself within the map in addition to being able to exploit map-information.
  
  This is the problem that is addressed by \cite{MiPaViICLR2017} with
        limited success.  They evaluate this case on two maps and report
        \LatencyOneGtOne{} to be greater than 1 in one of the two maps.
        We evaluate the same metric on ten randomized maps.
        \ditem{Random goal, random spawn, and random map} We believe
        that any proposed algorithms on end-to-end navigation problems,
        should be evaluated on unseen maps.  To our knowledge this is
        the first paper to do so in the case of deep reinforcement
        learning based navigation methods while reporting exploitation
        related metrics.  We train agents to simultaneously learn to
        explore 1000 maps and test them on 100 unseen maps. The relevant
        results can be found in Fig~\ref{fig:num-training-maps} and
        discussed in Section~\ref{sec:analysis}.
\end{description}

The comparative evaluation of the different the stages of this benchmark
are shown in Fig~\ref{fig:latency-goal-reward} and expanded upon in the
Section~\ref{sec:analysis}.

%% file: analysis.tex
We analyze the results obtained from evaluating the \NavAiiiCDiDiiL{} on our proposed experimental suite along with qualitative and quantitative additional studies.

\begin{figure}%
  \includegraphics[width=\linewidth]{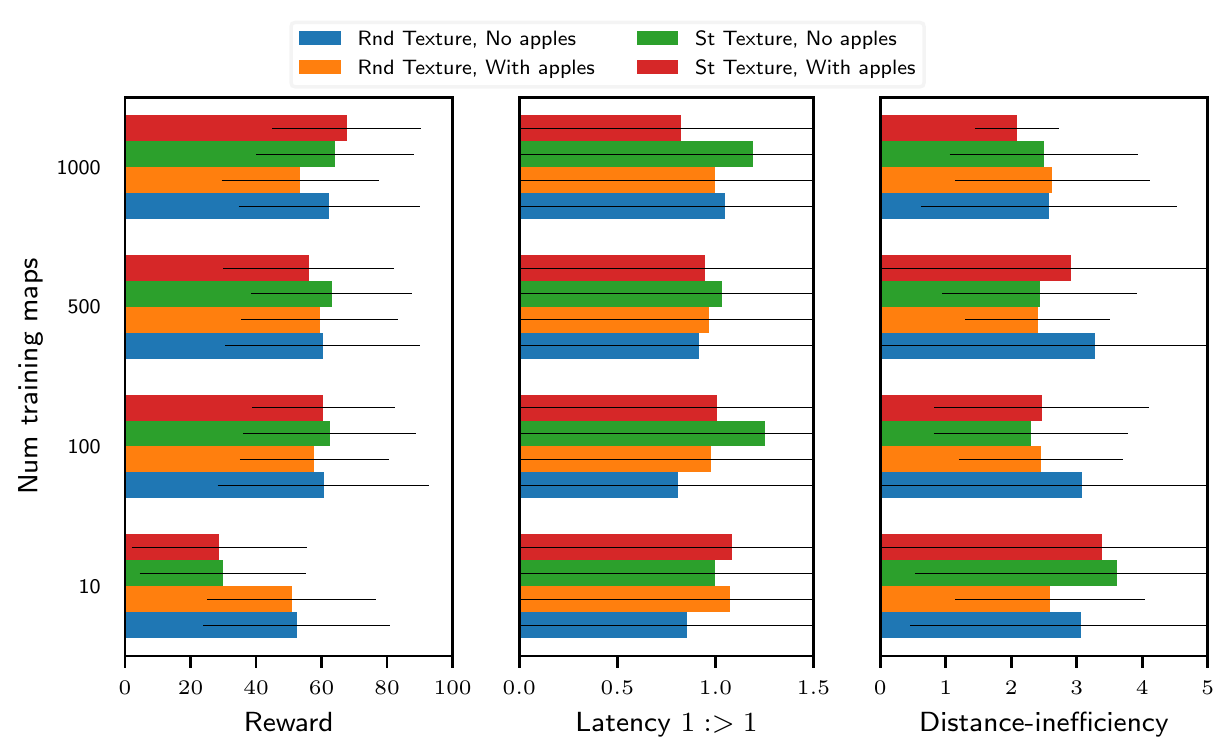}%
  \caption{Plots showing the effect of number of training maps with
    random texture (Rnd Texture) and presence of apples (With apples),
    when evaluated on unseen maps. We note that the difference
    between metrics is negligible compared to the standard deviation
    of the metrics. Hence we say that the algorithm is robust to the variation of textures and to the removal of apples.
    Note that the mean rewards increase when the number of training maps increases from  10 to 100 maps, but that rewards remain constant thereafter despite further increases in the number of training maps. }
  \label{fig:num-training-maps}
\end{figure}

\subsection{Static Map Experiments}

In this section we discuss the results for experiments as discussed in Section~\ref{sec:navtasks} over the ten randomly chosen maps. The results are in Fig~\ref{fig:latency-goal-reward}.
The experiment titled Random goal, random spawn, random maze is trained on 1000 maps but tested on the same ten maps for comparability.

\setcounter{Benchmark}{0}
\begin{description}
\ditem{Static goal, static spawn, static maze}
For this case, the reward is consistently high, and \DistanceInefficiency{} is close to 1 with small standard deviations implying the path chosen is the shortest available.
Please note that \LatencyOneGtOne{} should be close to 1 for the static
goal case because the goal location is learned  at training time.

\ditem{Static goal, random spawn, static map}
Again, note that \DistanceInefficiency{} is close to 1 implying that when the goal is found, the shortest path is traversed.
This is because the agent can learn the an optimal policy for the
shortest path to the known goal from any location in the maze at training time.

\ditem{Random goal, static spawn, static map} In this case, the mean
of the \LatencyOneGtOne{} is more than 1 showing that in general the
        agent is able to exploit map information.  However, as confirmed
        by the \DistanceInefficiency{} metric which is larger than one,
        that agents do not necessarily traverse the shortest path to the
        goal when it is found. 

\ditem{Random goal, Random spawn, static map} Similar to the
previous experiment, the \LatencyOneGtOne{} metric and the
\DistanceInefficiency{} metric is larger than 1 implying that while map
exploitation is taking place, the paths traversed to the goal are not
optimal.

\ditem{Random goal, Random spawn, Random map} For this experiment,
agents trained on a 1000 maps are tested individually on the 10 chosen
maps that are a subset of the 1000 maps.  The \LatencyOneGtOne{} is
close to 1 implying that map-exploitation is not taking place.  The
large \DistanceInefficiency{} numbers confirm this statement.  
\end{description}

\subsection{Random Map Experiments}
\label{sec:effect-apples-texture}

The following set of experiments are evaluated on 100 maps that are disjoint from the set of 1000 training maps. 

\subsubsection{Evaluation on unseen maps}
The results for training on subsets of 1000 maps, and testing on 100 unseen maps
are shown in Fig~\ref{fig:num-training-maps}.  We observe that there is
a jump of average reward and average goal hits when the number of
training maps is increased from 10 to 100 but no significant increase
when the number of training maps are increased from 100 to 500 to 1000.
We hypothesize that this is due to the fact that the algorithm learns an
average-exploration strategy which is learned with enough variation over
100 maps and training on additional maps does not add to it.

The \LatencyOneGtOne{} and
\DistanceInefficiency{} metrics confirm that no measurable map-exploitation is
taking place.  We present, qualitative results in
Sec.~\ref{sec:qualitative-simple} on very simple maps to show that the
agents trained on 1000 maps are only randomly exploring the maze rather
than utilizing any form of shortest path planning.

\subsubsection{Effect of apples and texture}
We evaluate the effect of apples and textures during evaluation time in
Fig~\ref{fig:num-training-maps}.  We train the algorithm on randomly
chosen training maps with random textures and evaluate them on maps with
and without random textures and also with and without apples.  We find
no significant changes across our different metrics showcasing that our
training mechanism allows for creation of agent's that are robust to the
presence or absence of textures and apples.

\subsubsection{Qualitative evaluation on simple maps}
\label{sec:qualitative-simple}
To evaluate what strategies the algorithm is employing to reach the
goal we evaluate the algorithm on very simple maps where there are only two
paths to reach the goal. The qualitative results for the evaluation are shown
in Fig~\ref{fig:planning-qualitative}.

\begin{description}
\item[Square map]
A Square map (Fig~\ref{fig:planning-qualitative}) is the simplest possible map with two paths to the goal.
We evaluate the algorithm trained on 1000 random maps on this map.
We observe that the agent greedily moves in the direction of
initialization.
We compute the percentage of times the agent takes the shortest path over a trial of 100 episodes.
We find the agent takes the shortest path only $50.4$\% ($\pm 12.8$\%) of the times which is no better than random.

%
\item[Goal map]
The goal map (Fig~\ref{fig:planning-qualitative}) provides a decision
point independent of the initial orientation. The shortest path is
chosen $42.6$\% ($\pm 35.1$\%) of the times which is again close to
random within error bounds.
\end{description}

These experiments show that the algorithm, even when trained on 1000
maps, does not generalize to these very simple maps highlighting how the
learned navigational strategy is unable to exploit map information

\input{fig-planning-qualitative}

\subsubsection{Attention heat maps}
\begin{figure}
\includegraphics[width=\columnwidth,trim=0 0 1008pt 336pt,clip]{training-09x09-0127-on-0127.png}\vspace{1ex}\\
\includegraphics[width=\columnwidth,trim=0 0 1008pt 336pt,clip]{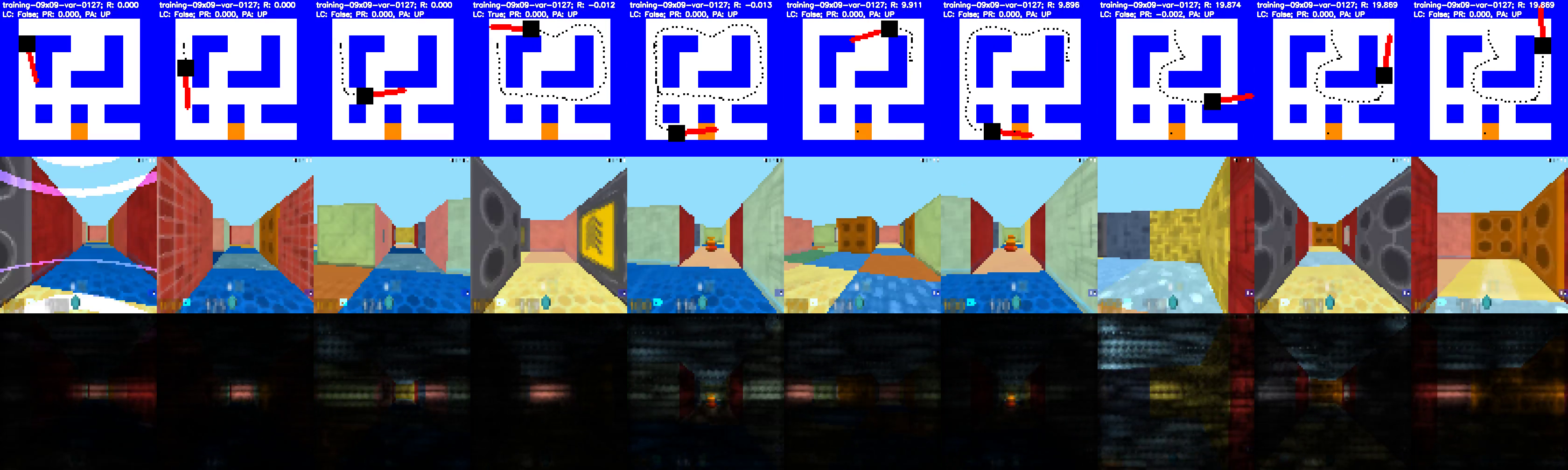}%
\caption{Visualizing attention for two sequences. The top two rows
    show the sequence when the model is trained on and evaluated on the
    same map. The bottom two rows show the sequence for a model trained
    on 1000 different maps and evaluated on one of those maps chosen at random.
    We observe that in
    both cases the attention is only on a few pixels in the center for
    the majority of the episode.}
\label{fig:attention}
\end{figure}

To qualitatively assess the visual cues used by agents in course of
their navigation, we use the normalized sum of absolute gradient of the
loss with respect to the input image as a proxy for attention in the
image.  The gradients are normalized for each image so that the maximum
gradient is one. The attention values are then used as a soft mask on
the image to create the visualization as shown in
Fig~\ref{fig:attention}

We observe that for most of the episode the attention narrows down to a
thin band in the center of the image. We hypothesize that this narrowing
down of the attention band highlights how the information required
for navigation can be found within this central band.
In future work, we will evaluate the performance of training agents using only this central band to see whether the resultant reward curves and metric scores are similar to that of the originals.

%% file: fig-planning-qualitative.tex
\begin{figure}[h]
\def\vertspace{1ex}
\rotatebox{90}{\hspace{1ex}\tiny Square map}%
\includegraphics[width=0.98\columnwidth,trim=0 672pt 1344pt 0,clip]{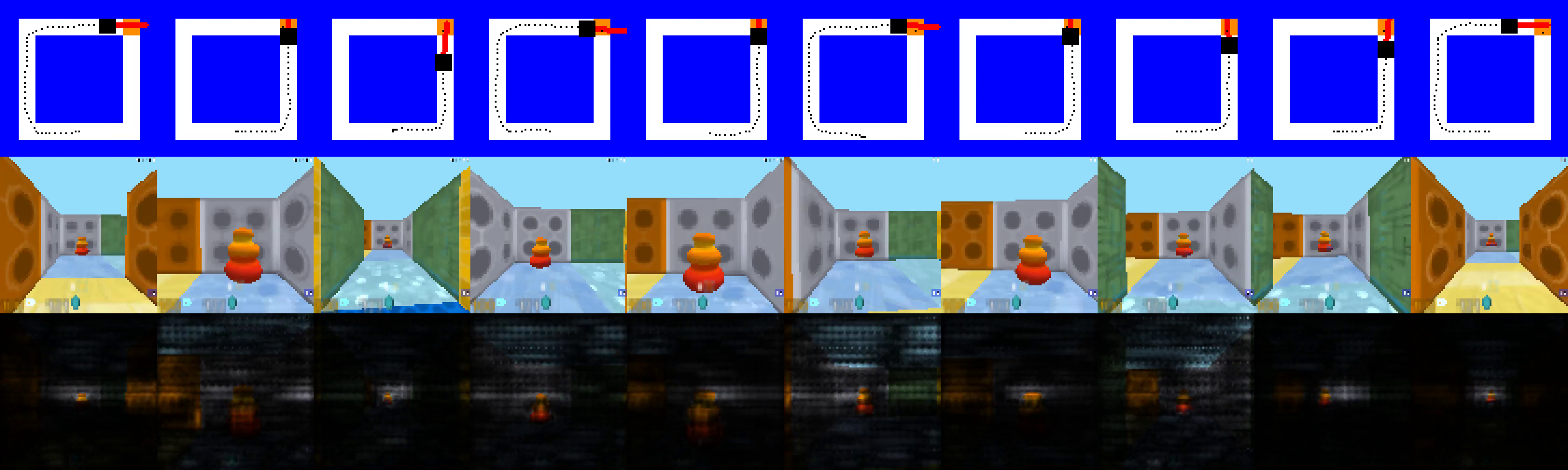}%
\vspace{\vertspace}
\rotatebox{90}{\hspace{2ex}\tiny Goal map}%
\includegraphics[width=0.98\columnwidth,trim=0 672pt 1004pt 0,clip]{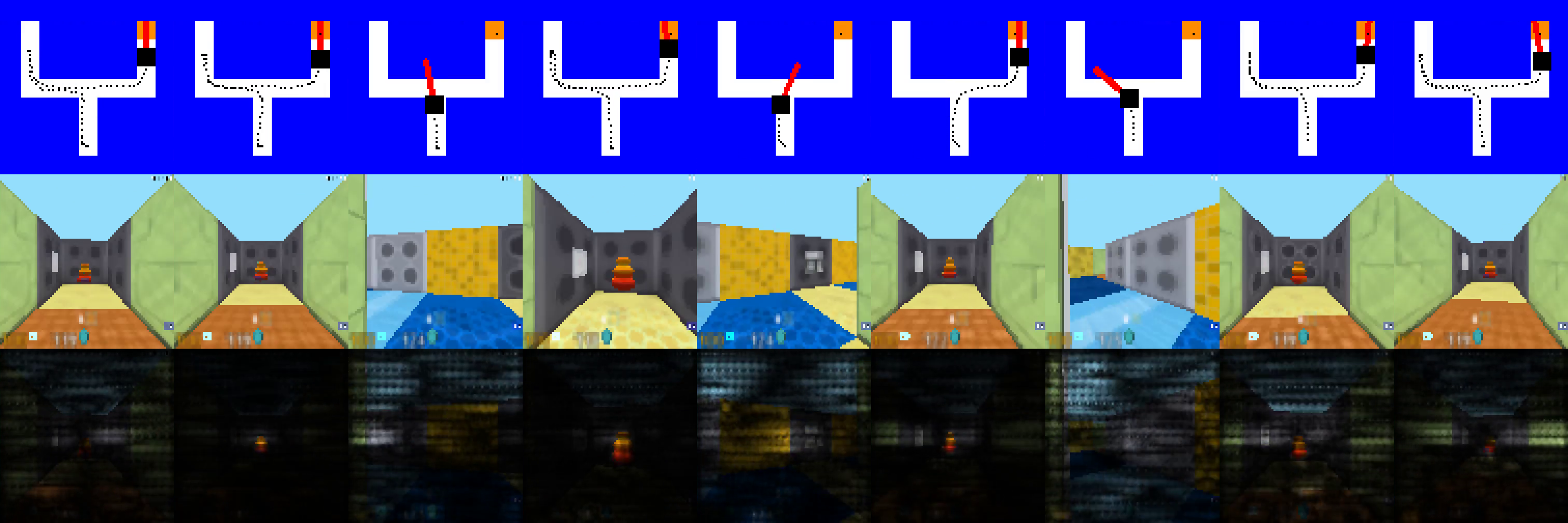}%
\caption{Snapshots of the path taken by the agent to reach the goal in a
    single episode when the model is trained on 1000 maps and evaluated
    on the Square and Goal maps.
  The top row shows an evaluation example on the Square map, where the agent takes the shortest path 3/6 times. 
  The bottom row shows an evaluation example on the Goal map, where the agent takes the shortest path 1/4 times.
  When averaged over 100 episodes, the agent performs no better than random as the shortest path is taken  $50.4$\% ($\pm 12.8$\%) and $42.6$\% ($\pm 35.1$\%) of the time for the Square map and the Goal map, respectively.
}
\label{fig:planning-qualitative}
\end{figure}

%% file: conclusion.tex
In this work, we present a systematic suite of experiments to
analyze DRL-based navigation algorithms.
We build upon \citet{MiPaViICLR2017}'s \NavAiiiCDiDiiL{} method for analysis.
We ask whether DRL algorithms are able to inherently gather and exploit environmental information over the course of the navigation of environments.
Our experiments show that the algorithm is able to exploit
map-information for navigation when trained and tested on same map,
yet unable to do so when trained and tested on different sets of maps.
Even when tested and trained on the same map, the DRL algorithms fail to find the shortest path to the destination location when it is randomly chosen.
These observations suggest that there is much more research to be done in DRL algorithms before they can compete with classical navigation techniques.

We believe that such rigorous investigation of DRL based navigation
methods is imperative to the field moving forward, especially given the
black-box nature of deep learning methods.  Recent work analyzing
neural networks has shown how deep learning-based can be fooled
in object detection \citep{NgYoClCVPR2015} and Atari games
\citep{kansky2017schema}
Such levels of sensitivity motivate exactly why it is important to analyze DRL
navigation methods across a wide variety of experiments: we need to
comprehensively understand both their strengths and limitations. As
such, we open source our experimental suite to serve as a framework
for the comparison of future DRL navigation studies.



%% file: main.bbl
\begin{thebibliography}{}

\bibitem[\protect\citeauthoryear{Aycock}{2016}]{Aycock2016}
John Aycock.
\newblock {\em Procedural Content Generation}, page 135.
\newblock Springer International Publishing, Cham, 2016.

\bibitem[\protect\citeauthoryear{Beattie \bgroup \em et al.\egroup
  }{2016}]{BeLeTeARXIV2016}
Charles Beattie, Joel~Z Leibo, Denis Teplyashin, Tom Ward, Marcus Wainwright,
  Heinrich K{\"u}ttler, Andrew Lefrancq, Simon Green, V{\'\i}ctor Vald{\'e}s,
  Amir Sadik, et~al.
\newblock Deepmind lab.
\newblock {\em arXiv preprint arXiv:1612.03801}, 2016.

\bibitem[\protect\citeauthoryear{Chaplot \bgroup \em et al.\egroup
  }{2016}]{ChLaSaNIPS2016}
Devendra~Singh Chaplot, Guillaume Lample, Kanthashree~Mysore Sathyendra, and
  Ruslan Salakhutdinov.
\newblock Transfer deep reinforcement learning in 3d environments: An empirical
  study.
\newblock 2016.

\bibitem[\protect\citeauthoryear{Elfes}{1989}]{ElCOMPUTER1980}
Alberto Elfes.
\newblock Using occupancy grids for mobile robot perception and navigation.
\newblock {\em Computer}, 22(6):46--57, 1989.

\bibitem[\protect\citeauthoryear{Gupta \bgroup \em et al.\egroup
  }{2017}]{gupta2017cognitive}
Saurabh Gupta, James Davidson, Sergey Levine, Rahul Sukthankar, and Jitendra
  Malik.
\newblock Cognitive mapping and planning for visual navigation.
\newblock In {\em The IEEE Conference on Computer Vision and Pattern
  Recognition (CVPR)}, July 2017.

\bibitem[\protect\citeauthoryear{Hausknecht and
  Stone}{2015}]{hausknecht2015deep}
Matthew Hausknecht and Peter Stone.
\newblock Deep recurrent q-learning for partially observable mdps.
\newblock {\em CoRR, abs/1507.06527}, 2015.

\bibitem[\protect\citeauthoryear{Jaderberg \bgroup \em et al.\egroup
  }{2016}]{jaderberg2016reinforcement}
Max Jaderberg, Volodymyr Mnih, Wojciech~Marian Czarnecki, Tom Schaul, Joel~Z
  Leibo, David Silver, and Koray Kavukcuoglu.
\newblock Reinforcement learning with unsupervised auxiliary tasks.
\newblock {\em arXiv preprint arXiv:1611.05397}, 2016.

\bibitem[\protect\citeauthoryear{Kansky \bgroup \em et al.\egroup
  }{2017}]{kansky2017schema}
Ken Kansky, Tom Silver, David~A M{\'e}ly, Mohamed Eldawy, Miguel
  L{\'a}zaro-Gredilla, Xinghua Lou, Nimrod Dorfman, Szymon Sidor, Scott
  Phoenix, and Dileep George.
\newblock Schema networks: Zero-shot transfer with a generative causal model of
  intuitive physics.
\newblock {\em arXiv preprint arXiv:1706.04317}, 2017.

\bibitem[\protect\citeauthoryear{Kuipers}{1978}]{KuCOGSCI1978}
Benjamin Kuipers.
\newblock Modeling spatial knowledge.
\newblock {\em Cognitive science}, 2(2):129--153, 1978.

\bibitem[\protect\citeauthoryear{Kulkarni \bgroup \em et al.\egroup
  }{2016}]{KuSaGaAPA2016}
Tejas~D Kulkarni, Ardavan Saeedi, Simanta Gautam, and Samuel~J Gershman.
\newblock Deep successor reinforcement learning.
\newblock {\em arXiv preprint arXiv:1606.02396}, 2016.

\bibitem[\protect\citeauthoryear{Levine \bgroup \em et al.\egroup
  }{2017}]{LePaKrISER2017}
Sergey Levine, Peter Pastor, Alex Krizhevsky, and Deirdre Quillen.
\newblock {\em Learning Hand-Eye Coordination for Robotic Grasping with
  Large-Scale Data Collection}, pages 173--184.
\newblock Springer International Publishing, Cham, 2017.

\bibitem[\protect\citeauthoryear{Mirowski \bgroup \em et al.\egroup
  }{2017}]{MiPaViICLR2017}
Piotr Mirowski, Razvan Pascanu, Fabio Viola, Hubert Soyer, Andrew~J. Ballard,
  Andrea Banino, Misha Denil, Ross Goroshin, Laurent Sifre, Koray Kavukcuoglu,
  Dharshan Kumaran, and Raia Hadsell.
\newblock Learning to navigate in complex environments.
\newblock 2017.

\bibitem[\protect\citeauthoryear{Mnih \bgroup \em et al.\egroup
  }{2013}]{MnKaSiNIPSDLW2013}
Volodymyr Mnih, Koray Kavukcuoglu, David Silver, Alex Graves, Ioannis
  Antonoglou, Daan Wierstra, and Martin Riedmiller.
\newblock Playing atari with deep reinforcement learning.
\newblock In {\em NIPS Deep Learning Workshop}. NIPS, 2013.

\bibitem[\protect\citeauthoryear{Mnih \bgroup \em et al.\egroup
  }{2015}]{MnKaSiNATURE2015}
Volodymyr Mnih, Koray Kavukcuoglu, David Silver, Andrei~A Rusu, Joel Veness,
  Marc~G Bellemare, Alex Graves, Martin Riedmiller, Andreas~K Fidjeland, Georg
  Ostrovski, et~al.
\newblock Human-level control through deep reinforcement learning.
\newblock {\em Nature}, 518(7540):529--533, 2015.

\bibitem[\protect\citeauthoryear{Mnih \bgroup \em et al.\egroup
  }{2016}]{MnBaMiICML2016}
Volodymyr Mnih, Adria~Puigdomenech Badia, Mehdi Mirza, Alex Graves, Timothy
  Lillicrap, Tim Harley, David Silver, and Koray Kavukcuoglu.
\newblock Asynchronous methods for deep reinforcement learning.
\newblock In {\em International Conference on Machine Learning}, pages
  1928--1937, 2016.

\bibitem[\protect\citeauthoryear{Nguyen \bgroup \em et al.\egroup
  }{2015}]{NgYoClCVPR2015}
Anh Nguyen, Jason Yosinski, and Jeff Clune.
\newblock Deep neural networks are easily fooled: High confidence predictions
  for unrecognizable images.
\newblock In {\em 2015 IEEE Conference on Computer Vision and Pattern
  Recognition (CVPR)}, pages 427--436, June 2015.

\bibitem[\protect\citeauthoryear{Oh \bgroup \em et al.\egroup
  }{2016}]{OhChSiICML2016}
Junhyuk Oh, V.~Chockalingam, S.~Singh, and H.~Lee.
\newblock Control of memory, active perception, and action in minecraft.
\newblock In {\em International Conference on Machine Learning}, 2016.

\bibitem[\protect\citeauthoryear{Parisotto and
  Salakhutdinov}{2017}]{parisotto2017neural}
Emilio Parisotto and Ruslan Salakhutdinov.
\newblock Neural map: Structured memory for deep reinforcement learning.
\newblock {\em arXiv preprint arXiv:1702.08360}, 2017.

\bibitem[\protect\citeauthoryear{Silver \bgroup \em et al.\egroup
  }{2016}]{SiHuMaNATURE2016}
David Silver, Aja Huang, Chris~J Maddison, Arthur Guez, Laurent Sifre, George
  Van Den~Driessche, Julian Schrittwieser, Ioannis Antonoglou, Veda
  Panneershelvam, Marc Lanctot, et~al.
\newblock Mastering the game of go with deep neural networks and tree search.
\newblock {\em Nature}, 529(7587):484--489, 2016.

\bibitem[\protect\citeauthoryear{Smith and Cheeseman}{1986}]{SmChIJRR1986}
Randall~C Smith and Peter Cheeseman.
\newblock On the representation and estimation of spatial uncertainty.
\newblock {\em The international journal of Robotics Research}, 5(4):56--68,
  1986.

\bibitem[\protect\citeauthoryear{Sutton and Barto}{1998}]{SuBaBOOK1998}
Richard~S Sutton and Andrew~G Barto.
\newblock {\em Reinforcement learning: An introduction}, volume~1.
\newblock MIT press Cambridge, 1998.

\bibitem[\protect\citeauthoryear{Zhu \bgroup \em et al.\egroup
  }{2017}]{ZhMoKoICRA2017}
Yuke Zhu, Roozbeh Mottaghi, Eric Kolve, Joseph~J Lim, Abhinav Gupta,
  Li~Fei-Fei, and Ali Farhadi.
\newblock Target-driven visual navigation in indoor scenes using deep
  reinforcement learning.
\newblock In {\em Robotics and Automation (ICRA), 2017 IEEE International
  Conference on}, pages 3357--3364. IEEE, 2017.

\end{thebibliography}
